\documentclass[runningheads]{llncs}
\usepackage{graphicx}
\usepackage{multirow}
\usepackage[table,xcdraw]{xcolor}
\usepackage{tikz}
\usepackage{comment}
\usepackage{amsmath,amssymb} % define this before the line numbering.
\usepackage{color}

\usepackage[pagebackref,breaklinks,colorlinks]{hyperref}
\newcommand{\etal}{\textit{et al}. }
\newcommand{\ie}{\textit{i}.\textit{e}. }
\newcommand{\eg}{\textit{e}.\textit{g}. }
\definecolor{demphcolor}{RGB}{144,144,144}

\usepackage{orcidlink}

\newcommand\blfootnote[1]{%
\begingroup
\renewcommand\thefootnote{}\footnote{#1}%
\addtocounter{footnote}{-1}%
\endgroup
}

\usepackage[accsupp]{axessibility}  % Improves PDF readability for those with disabilities.

\begin{document}
% \renewcommand\thelinenumber{\color[rgb]{0.2,0.5,0.8}\normalfont\sffamily\scriptsize\arabic{linenumber}\color[rgb]{0,0,0}}
% \renewcommand\makeLineNumber {\hss\thelinenumber\ \hspace{6mm} \rlap{\hskip\textwidth\ \hspace{6.5mm}\thelinenumber}}
% \linenumbers
\pagestyle{headings}
\mainmatter
\def\ECCVSubNumber{6671}  % Insert your submission number here

\title{3D Interacting Hand Pose Estimation by Hand De-occlusion and Removal}

% INITIAL SUBMISSION 
\begin{comment}
\titlerunning{ECCV-22 submission ID \ECCVSubNumber} 
\authorrunning{ECCV-22 submission ID \ECCVSubNumber} 
\author{Anonymous ECCV submission}
\institute{Paper ID \ECCVSubNumber}
\end{comment}
%******************

% CAMERA READY SUBMISSION
% \begin{comment}
\titlerunning{Hand De-occlusion and Removal}
% If the paper title is too long for the running head, you can set
% an abbreviated paper title here
%

\author{
    Hao Meng\inst{1,3*}\orcidlink{0000-0001-8136-0080} \and
    Sheng Jin\inst{2,3*}\orcidlink{0000-0001-5736-7434} \and
	Wentao Liu\inst{3,4}\orcidlink{0000-0001-6587-9878} \and
	Chen Qian\inst{3}\orcidlink{0000-0002-8761-5563} \\ Mengxiang Lin\inst{1} \and Wanli Ouyang\inst{4,5}\orcidlink{0000-0002-9163-2761} \and  Ping Luo\inst{2}\orcidlink{0000-0002-6685-7950}
	 \\[.21cm]
	$^{1}$ Beihang University \quad
	$^{2}$ The University of Hong Kong \quad
	$^{3}$ SenseTime Research and Tetras.AI \quad
	$^{4}$ Shanghai AI Lab \quad
	$^{5}$ The University of Sydney \\
	\tt\small hao\_meng@163.com, \{jinsheng, liuwentao, qianchen\}@tetras.ai \\
	\tt\small linmx@buaa.edu.cn, wanli.ouyang@sydney.edu.au, pluo@cs.hku.hk
	}
\authorrunning{H. Meng et al.}
% First names are abbreviated in the running head.
% If there are more than two authors, 'et al.' is used.
%
\institute{}

% \end{comment}
%******************
\maketitle

\begin{abstract}

Estimating 3D interacting hand pose from a single RGB image is essential for understanding human actions. Unlike most previous works that directly predict the 3D poses of two interacting hands simultaneously, we propose to decompose the challenging interacting hand pose estimation task and estimate the pose of each hand separately. In this way, it is straightforward to take advantage of the latest research progress on the single-hand pose estimation system. However, hand pose estimation in interacting scenarios is very challenging, due to (1) severe hand-hand occlusion and (2) ambiguity caused by the homogeneous appearance of hands. To tackle these two challenges, we propose a novel Hand De-occlusion and Removal (HDR) framework to perform hand de-occlusion and distractor removal. We also propose the first large-scale synthetic amodal hand dataset, termed Amodal InterHand Dataset (AIH), to facilitate model training and promote the development of the related research. Experiments show that the proposed method significantly outperforms previous state-of-the-art interacting hand pose estimation approaches. Codes and data are available at \url{https://github.com/MengHao666/HDR}.


\blfootnote{* Equal Contribution.}

\end{abstract}

\section{Introduction}

\begin{figure*}[t]
\begin{center}
   \includegraphics[width=0.9\linewidth]{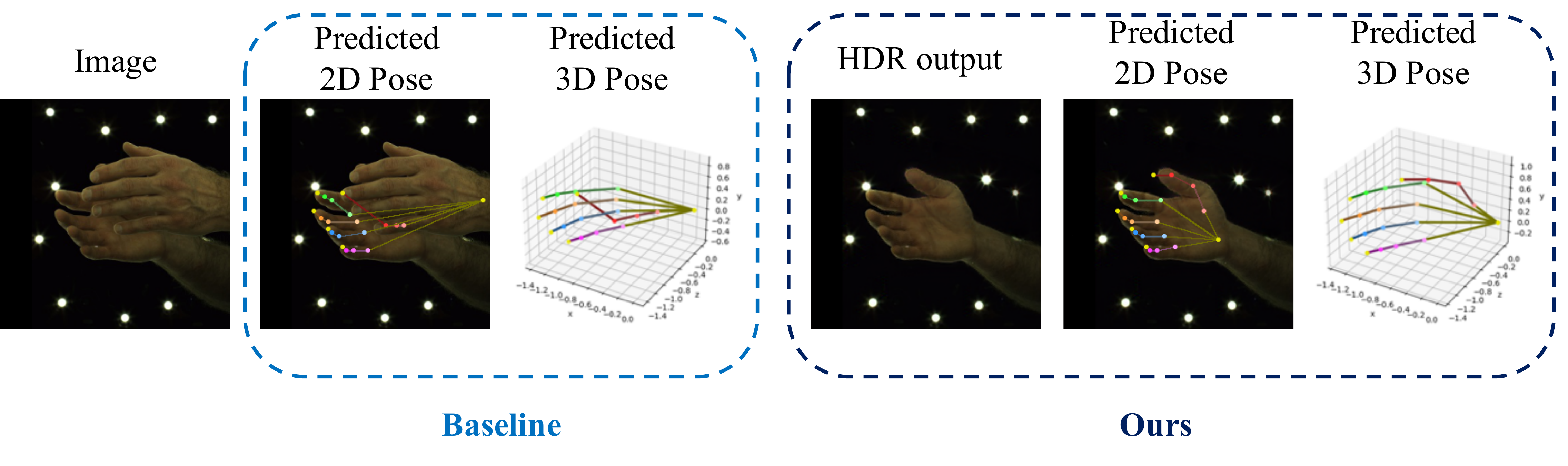}
\end{center}
   \caption{State-of-the-art hand pose estimation models often struggle to estimate the 3D poses of interacting hands, due to severe hand-hand occlusion and appearance ambiguity of two hands. In this example, we observe erroneous pose estimation of the occluded part and significant uncertainty between the left and right wrist. Our HDR framework tackles these two challenges via hand de-occlusion (recovering the appearance content of the occluded part) and removal (removing the other distracting hand). It transforms the challenging interacting hand image into a simple single-hand image, which can be easily handled by the hand pose estimator.
   }
\label{fig:motivation}
\end{figure*}

Estimating the 3D hand pose from a monocular RGB image is critical in many real-world applications, \eg human-computer interaction, augmented and virtual reality (AR/VR), and sign language recognition. Although significant progress has been made for single-hand pose estimation, analysis of hand-hand interactions remains challenging. Estimating 3D interacting hand pose from a single RGB image has attracted increasing research attention in recent years.

In this paper, we propose to decompose the challenging interacting hand pose estimation task, and predict the pose of the left and the right hand separately. However, solving single-hand pose estimation in close two-hand interaction cases is non-trivial, because of two major challenges. One of the main challenges is the severe hand-hand occlusion. Considering hands in close interactions, the occlusion patterns are complex. Many areas of the target hand can be occluded, making it very challenging to infer the pose of the invisible parts. Another challenge is that the homogeneous and self-similar appearance of hands (\ie the left and the right hand) may cause ambiguities. And the hand pose estimator may be confused by the other visually similar distracting hand.

To tackle these challenges, we propose a simple yet effective Hand De-occlusion and Removal (HDR) framework. Specifically, our HDR framework comprises three parts, Hand Amodal Segmentation Module (HASM), Hand De-occlusion and Removal Module (HDRM), and the Single Hand Pose Estimator (SHPE). 
HAS segments both the complete (amodal) and visible parts for both two hands. The resulting segmentation masks not only contain information to localize the rough position of the two hands, but also provide cues for subsequent de-occlusion and removal process by HDRM. \emph{De-occlusion} targets at predicting the appearance content of the occluded part. \emph{Removal} targets at removing the distracting part in the image. In our case, when estimating the pose of the right hand, the left hand becomes the distracting part and should be removed. As shown in Fig.~\ref{fig:motivation}, recent state-of-the-art hand pose estimation methods suffer from severe hand-hand occlusion and the homogeneous appearance of two hands, resulting in inferior pose estimation results. Thanks to our proposed HDR framework, we can transform the challenging scenario of hand-hand interactions into a common single-hand scenario, which can be easily handled by an off-the-shelf SHPE. 

However, to the best of our knowledge, there exist no datasets that contain both the amodal segmentation and appearance content ground-truths of interactive hands. To fill in this blank, we synthetically generate a large-scale Amodal InterHand dataset, namely AIH dataset. The dataset contains over 3 million interacting hand images along with ground-truth amodal and modal segmentation, de-occlusion and removal ground-truths. The dataset consists of two parts, \ie AIH\_Syn and AIH\_Render. AIH\_Syn is obtained by simple random copy-and-paste. It retains detailed and realistic appearance information. However, it may generate implausible interacting hand poses that violate the biomechanical structure of the human body. AIH\_Render is generated by rendering the textured 3D interacting hand mesh to the image plane. The inter-dependencies between two hands are fully considered to avoid physically implausible
configurations, \eg intersecting fingers. However, it may suffer from the appearance gap because the rendered texture is synthetic. Combining the advantages of both, we make a large-scale 3D hand-hand interaction dataset with large pose and appearance variety. We empirically validate the effectiveness of the synthetic dataset through extensive experiments. We envision that our proposed dataset will foster the development of the related research, \eg interacting hand pose estimation, amodal or modal instance segmentation, de-occlusion, etc.

Our proposed Hand De-occlusion and Removal (HDR) framework is simple, flexible, and effective. Extensive experiments on the well-known InterHand2.6M benchmark~\cite{moon2020interhand2} show that our method significantly outperforms the state-of-the-art 3D interacting hand pose estimation systems. Our framework builds upon the latest research progress of amodal segmentation~\cite{xie2021segformer}, de-occlusion~\cite{zhou2021human,zhan2020self,liu2018partialinpainting}, and 3D single-hand pose estimation~\cite{zhou2020monocular}. Note that, we do not perform complete comparisons with previous amodal segmentation, de-occlusion, and SHPE approaches. We also do not claim any algorithmic superiority concerning model architecture design. Because our aim is to propose a framework to solve the challenges of 3D interacting hand pose estimation. And designing powerful modules to improve the performance of amodal segmentation, de-occlusion, and SHPE is not the focus of this paper.

Our contributions are summarized as follows:

\begin{itemize}

\item We propose a novel Hand De-occlusion and Removal (HDR) framework to tackle the challenging task of 3D interacting hand pose estimation. 

\item We propose to explicitly handle the challenges of self-occlusion by hand de-occlusion and the homogeneous appearance ambiguity by distractor removal. To the best of our knowledge, we are the first to apply de-occlusion techniques to improve the downstream pose estimation accuracy.

\item We propose the first large-scale synthetic Amodal InterHand Dataset (AIH) to settle the task of hand de-occlusion and removal. We envision that AIH will foster the development of the related research.

\end{itemize}

\section{Related Work}

\subsection{Amodal Instance Segmentation and De-occlusion}

\textbf{Amodal Instance Segmentation.} Unlike modal instance segmentation, which aims at assigning labels to visible parts of instances, amodal instance segmentation targets at producing the amodal (integrated) masks of each object instance involving its occluded parts. Li and Malik~\cite{li2016amodal} proposed the first amodal instance segmentation model which iteratively expands the bounding boxes and recomputes the heatmaps. Zhu~\etal~\cite{zhu2017semantic} proposed COCOA dataset for amodal instance segmentation and presented AmodalMask model as the baseline. Zhan~\etal~\cite{zhan2020self} propose a method to reason about the underlying occlusion ordering and recover the invisible parts in a self-supervised manner. 

\textbf{De-occlusion.} De-occlusion aims at recovering the appearance content of the invisible occluded parts. SeGAN~\cite{ehsani2018segan} adopts a residual network based model for mask completion and inferring the appearance of the invisible parts of indoor objects. Yan~\etal~\cite{yan2019visualizing} presented an iterative multi-task framework for amodal mask completion and de-occlusion of vehicles. Zhou~\etal~\cite{zhou2021human} built upon a well-known inpainting approach~\cite{liu2018partialinpainting} and proposed to reason about the occluded regions and recover the appearance content of humans. Baek~\etal~\cite{baek2020weakly} presents a weakly-supervised method to adapt from hand-object domain to single hand-only domain. However, its image generation module and the pose estimator are deeply coupled together, limiting its generalization ability to adapt to different hand pose estimators and resulting in low-quality restored image. 

Our approach differs from previous works in three major aspects. First, previous works mostly focus on improving the quality of image content recovery, while we aim to improve the performance of the downstream task, \ie 3D interactive hand pose estimation. 
Second, compared with common rigid objects, recovering the appearance content of the interacting hands is more challenging because of larger pose variations, severe hand-hand occlusion, and self-similar appearance of hands and fingers. Third, besides de-occlusion, our proposed HDR framework also performs distracting hand removal to reduce the ambiguities caused by the homogeneous appearance of hands.

\subsection{Monocular RGB-based Hand Pose Estimation}

\textbf{Isolated hand pose estimation.} RGB-based single (isolated) hand pose estimation has made significant progress in the past few years. Zimmermann~\etal~\cite{zimmermann2017learning} introduced one of the first deep learning models to estimate hand poses from monocular RGB images. It first uses HandSegNet to localize hand regions, then uses PoseNet to estimate 2D hand poses, and finally maps 2D poses into 3D space. Iqbal~\etal~\cite{iqbal2018hand} proposed to encode hand joint locations with 2.5D heatmap representation to address the depth ambiguity problems and improve localization precision. Spurr~\etal~\cite{spurr2018cross} proposed a VAE-based generative model to regress 3D hand joint locations. Zhou~\etal~\cite{zhou2020monocular} proposed to fully exploit non-image MoCap data to improve model generalization and robustness. Recently, many works also attempt to estimate 3D hand meshes from monocular RGB images. Most of them~\cite{baek2019pushing,boukhayma20193d,yang2020bihand} are model-based, which train a convolutional neural network to estimate the MANO parameters~\cite{romero2022embodied}. Others are model-free, which directly regress 3D vertices of the human hand using mesh convolution~\cite{kulon2020weakly}, graph neural networks~\cite{Choi_2020_ECCV_Pose2Mesh}, or transformers~\cite{lin2021end}. 

\textbf{Interacting hand pose estimation.} Most works conduct interacting two-hand pose estimation by utilizing multi-view RGB images~\cite{ballan2012motion,han2020megatrack}, depth data~\cite{oikonomidis2012tracking,mueller2019real,tzionas2016capturing}, and tracking strategy~\cite{oikonomidis2012tracking,smith2020constraining,wang2020rgb2hands}. Only a few existing works have considered estimating 3D poses of two hands from a single RGB image, which is challenging due to severe occlusion and close interactions. Lin~\etal~\cite{lin2021two} employed a synthetic egocentric hand dataset to learn to estimate two-hand poses from a single RGB image. Moon \etal proposed a large-scale interacting hand dataset, termed InterHand2.6M dataset~\cite{moon2020interhand2}, and designed the InterNet model to predict 2.5D hand poses. 
Zhang~\etal~\cite{zhang2021interacting} designed a hand pose-aware attention module to address the self-similarity ambiguities and leveraged a context-aware cascaded refinement module to improve pose accuracy. Kim~\etal~\cite{kim2021end} introduced an end-to-end trainable framework to jointly perform interacting hand pose estimation. Rong~\etal~\cite{rong2021ihmr} presented a two-stage framework to generate precise 3D hand poses and meshes with minimal collisions from monocular single RGB images. Fan~\etal~\cite{fan2021digit} proposed DIGIT (DIsambiGuating hands in InTeraction) to explicitly leverage the per-pixel probabilities to reduce the ambiguities caused by self-similarity of hands. 

In this work, we empirically show that existing hand pose estimators often suffer from extreme self-occlusions and appearance ambiguity. To this end, we propose a novel Hand De-occlusion and Removal (HDR) framework to explicitly handle these two challenges, which significantly outperforms prior arts.

\section{Method}

\begin{figure*}[t]
\begin{center}
   \includegraphics[width=0.99\linewidth]{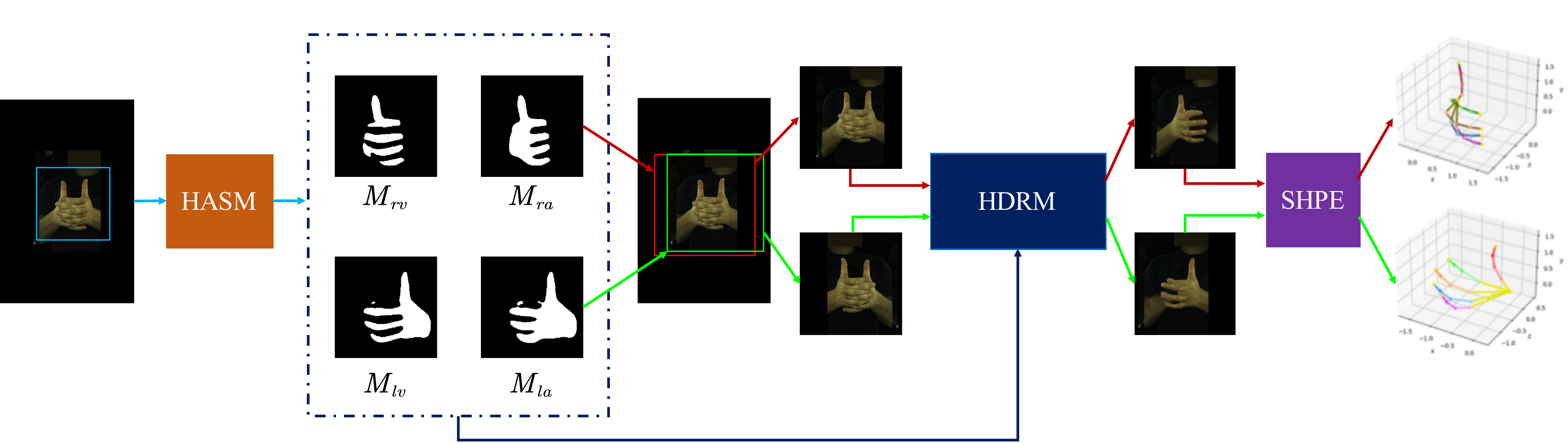}
\end{center}
   \caption{Illustration of our Hand De-occlusion and Removal (HDR) framework for the task of 3D interacting hand pose estimation. 
   We first employ \textbf{HASM} (Hand Amodal Segmentation Module) to segment the amodal and modal masks of the left and the right hand in the image. Given the predicted masks, we locate and crop the image patch centered at each hand. Then, for every cropped image, the \textbf{HDRM} (Hand De-occlusion and Removal Module) recovers the appearance content of the occluded part of one hand and removes the other distracting hand simultaneously. In this way, the interacting two-hand image is transformed into a single-hand image, and can be easily handled by \textbf{SHPE} (Single Hand Pose Estimation) to get the final 3D hand poses. 
   }
\label{fig:overview}
\end{figure*}

\subsection{Overview}

As shown in Fig.~\ref{fig:overview}, we propose a three-stage framework for interactive hand pose estimation. The first stage segments the complete and visible part for both two hands. The second stage recovers the RGB values of the occluded hand and the background behind the distracting hand at the same time. The third stage predicts the 3D pose of each hand separately. 

\subsection{Hand Amodal Segmentation Module (HASM)}
As shown in Fig.~\ref{fig:overview}, given an interacting two-hand image, we first obtain the modal and amodal masks of both hands using the Hand Amodal Segmentation Module (HASM). We simply adapt the off-the-shelf instance segmentation model, \ie SegFormer~\cite{xie2021segformer}, to fit in our two-hand amodal segmentation tasks. Specifically, we increase the number of decode heads from one to four to predict four kinds of segmentation masks, namely the right hand amodal mask $M_{ra}$, the right hand visible mask $M_{rv}$, the left hand amodal mask $M_{la}$ and the left hand visible mask $M_{lv}$. These segmentation masks contain (1) spatial localization information to roughly localize the left/right hand, and (2) rich cues about the occluded regions for de-occlusion and the distractor regions for removal.

We apply the binary cross entropy losses  
$\mathcal{L} _{BCE}\left( * \right)$ to supervise the segmentation model. The final segmentation loss functions are formulated as follows:

\begin{equation}
\begin{aligned}
\mathcal{L}_{HAS}&= \mathcal{L}_{BCE}\left( M_{ra}, M_{ra}^* \right) + \mathcal{L}_{BCE}\left( M_{lv}, M_{lv}^* \right) +\\
&\quad \ \mathcal{L} _{BCE}\left( M_{la}, M_{la}^* \right) +\mathcal{L} _{BCE}\left( M_{lv}, M_{lv}^* \right),
\end{aligned}
\label{loss_seg}
\end{equation}

\noindent where $M_{ra}$, $M_{lv}$, $M_{la}$, and $M_{lv}$ are predicted segmentation masks; $M_{ra}^*$, $M_{lv}^*$, $M_{la}^*$, and $M_{lv}^*$ are the corresponding ground-truth masks.

\subsection{Hand De-occlusion and Removal Module (HDRM)}

\begin{figure*}[t]
\begin{center}
   \includegraphics[width=0.99\linewidth]{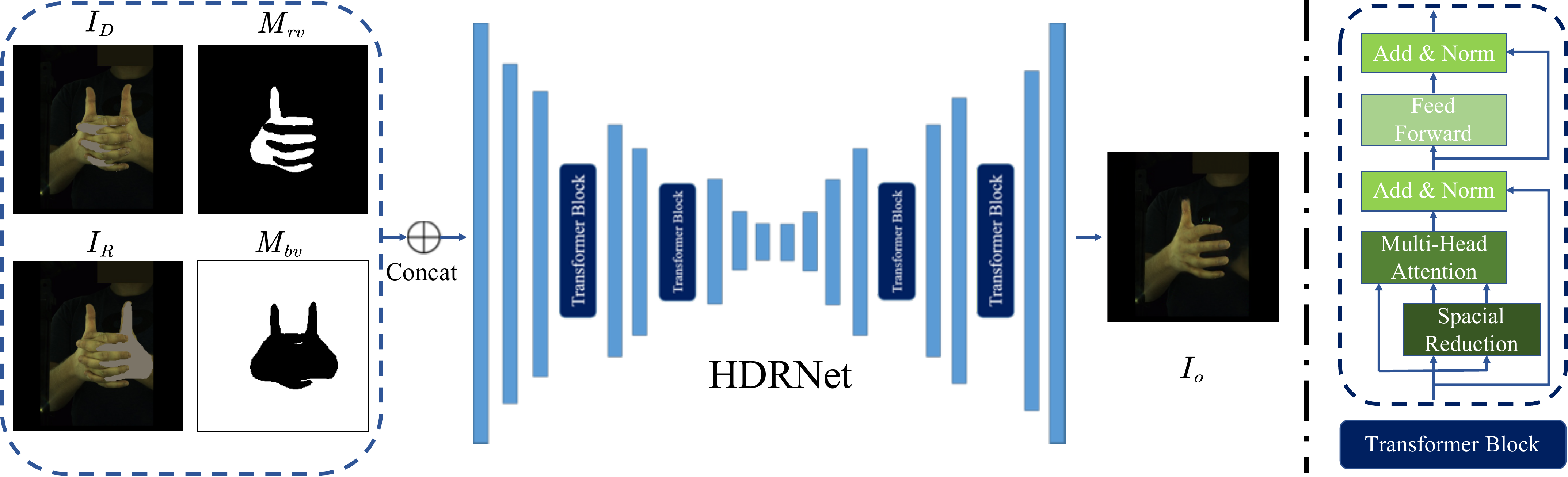}
\end{center}
   \caption{Illustration of the HDRNet. The input of HDRNet includes 4 kinds of data: (a) the image erased on occluded portion of the right hand $I_D$, (b) the modal mask of the right hand $M_{rv}$, (c) the image erased on redundant portion of the distracting hand $I_R$, and (d) the modal mask of background $M_{bv}$. HDRNet recovers the appearance content of the occluded parts and inpaints the distracting hand to avoid ambiguity.}
\label{fig:HDRNet}
\end{figure*}

Hand De-occlusion and Removal Module (HDRM) aims at transforming a previously challenging case of hand-hand interactions into a common single-hand case, which can be easily solved by an off-the-shelf single-hand pose estimator. Specifically, given amodal and modal masks, \emph{De-occlusion} is responsible for recovering the appearance content or RGB values of the occluded regions, while \emph{Removal} targets at inpainting the distracting regions in the image, reducing the ambiguities caused by the homogeneous appearance of two hands.

For clarity, in the following sections, we will focus on the right hand only and regard the left hand as the distractor. Note that, the left-hand centered image can be flipped horizontally before performing hand de-occlusion, removal, and pose estimation, thus following the same pipeline. 

As shown in Fig.~\ref{fig:overview}, for the right hand, we first use the amodal mask $M_{ra}$ to locate the right hand. Then we crop the original image and the segmentation masks at the center of the right hand. The newly cropped image and masks are denoted as $I_s^{crop}$, $M_{ra}^{crop}$, $M_{rv}^{crop}$, $M_{la}^{crop}$ and $M_{lv}^{crop}$ respectively. We will omit the superscript ${crop}$ in subsequent sections for simplicity.

We use $M_D$ to denote the region where the target hand is occluded by the other hand and $M_R$ to denote the region where the distracting hand occupies. They are computed as follows:

\begin{equation}
\begin{aligned}
& M_D=M_{ra}\cdot \left( 1-M_{rv} \right), \\
& M_R=(1-M_{ra})\cdot M_{lv}.
\end{aligned}
\label{loss_seg}
\end{equation}

$I_D$ and $I_R$ are the original image $I_s$ erased by the mask $M_D$ and $M_R$ respectively. They can inform the HDRNet where to focus and how to inpaint these two regions with partial convolution~\cite{liu2018partialinpainting}. In addition, the modal mask of the right hand $M_{rv}$ and the modal mask of the background $M_{bv}$ point out where the HDRNet can refer to for de-occlusion and removal respectively. 
Formally, $I_D$, $I_R$ and $M_{bv}$ are computed as follows:

\begin{equation}
\begin{aligned}
& I_D =I_s \cdot (1-M_{D}), \\
& I_R =I_s \cdot (1-M_{R}), \\
& M_{bv} =(1-M_{ra})\cdot(1-M_{la}).
\end{aligned}
\label{loss_seg}
\end{equation}

$I_D$, $M_{rv}$, $I_R$ and $M_{bv}$ are concatenated together as the input, as shown in Fig.~\ref{fig:HDRNet}. HDRNet then uses these data to recover the appearance content of the occluded parts and inpaints the distracting hand to avoid ambiguity. For model architecture choice, we follow~\cite{zhou2021human,zhan2020self} to adopt the network of Liu~\etal~\cite{liu2018partialinpainting} and further improve it by adding a few transformer blocks~\cite{wang2021pvtv2}. The transformer block enhances image feature interactions, enlarges the receptive fields, and focuses more on important image regions. Finally, the HDRNet outputs a recovered image $I_o$. We follow~\cite{zhou2021human} to employ an image discriminator~\cite{isola2017image} $D$ to enhance the image recovery quality through adversarial training. 
The loss function of HDRNet is as follows:

\begin{equation}
\begin{aligned}
\mathcal{L}_{HDR}= & \lambda_1 (\mathbb{E}_{I_o}[\mathrm{log}(1-D(I_o))] + \mathbb{E}_{I_o^*}[\mathrm{log}(D(I_o^*))]) + \\
& \lambda_2 \mathcal{L}_{\ell1}(I_o, I_o^*) + \lambda_3 \mathcal{L}_{prec}(I_o, I_o^*) + \lambda_4 \mathcal{L}_{style}(I_o, I_o^*),
\end{aligned}
\label{loss_HDR}
\end{equation}

\noindent where $\mathcal{L}_{prec}(*)$ denotes the perceptual loss~\cite{gatys2016image}, and $\mathcal{L}_{style}(*)$ denotes the style loss~\cite{liu2018partialinpainting}. $I_o$ is the recovered image, while $I_o^*$ is its corresponding ground truth. $\lambda_1$, $\lambda_2$, $\lambda_3$, and $\lambda_4$ are hyper-parameters to balance the losses.

\subsection{3D Single Hand Pose Estimation (SHPE)}

Our de-occlusion and removal framework can be applied to any off-the-shelf pose estimators. However, designing a more powerful hand pose estimation network architecture is not the focus of this paper. In this work, we choose the DetNet of MinimalHand~\cite{zhou2020monocular} as our baseline SHPE for its simplicity and good performance. MinimalHand~\cite{zhou2020monocular} comprises two modules, \ie DetNet and IKNet. DetNet predicts the 2D and 3D hand joint positions. IKNet then takes as input the predicted 3D hand joint positions and maps them to the joint angles. In our implementation, we simply discard the IKNet and re-train the DetNet on the InterHand2.6M dataset~\cite{moon2020interhand2}.
The loss function of SHPE is as follows:

\begin{equation}
\begin{aligned}
\mathcal{L}_{SHPE}= \mathcal{L}_{heat} + \mathcal{L}_{loc} + \mathcal{L}_{delta}+ \mathcal{L}_{reg},
\end{aligned}
\label{loss_SHPE}
\end{equation}

\noindent where $ \mathcal{L}_{heat} $ is the 2D heatmap loss. $\mathcal{L}_{loc}$ and $\mathcal{L}_{delta}$ are location map loss and delta map loss respectively. $\mathcal{L}_{reg}$ is a $\ell_2$ weight regularizer to avoid overfitting. Please refer to Zhou~\etal~\cite{zhou2020monocular} for more training details.

\section{Amodal InterHand (AIH) Dataset}

Existing amodal perception datasets~\cite{zhu2017semantic,qi2019amodal,hu2019sail} mostly focus on amodal segmentation of common objects (\eg, vehicles, buildings, and indoor objects). To the best of our knowledge, there exists no dataset that targets at amodal segmentation and appearance content recovery of interactive hands. To fill in this blank, we synthetically generate the first large-scale Amodal InterHand dataset, namely AIH dataset. We envision that the proposed dataset will boost the related research, \eg amodal perception, de-occlusion, and hand pose estimation.

\begin{figure*}[t]
\begin{center}
   \includegraphics[width=0.99\linewidth]{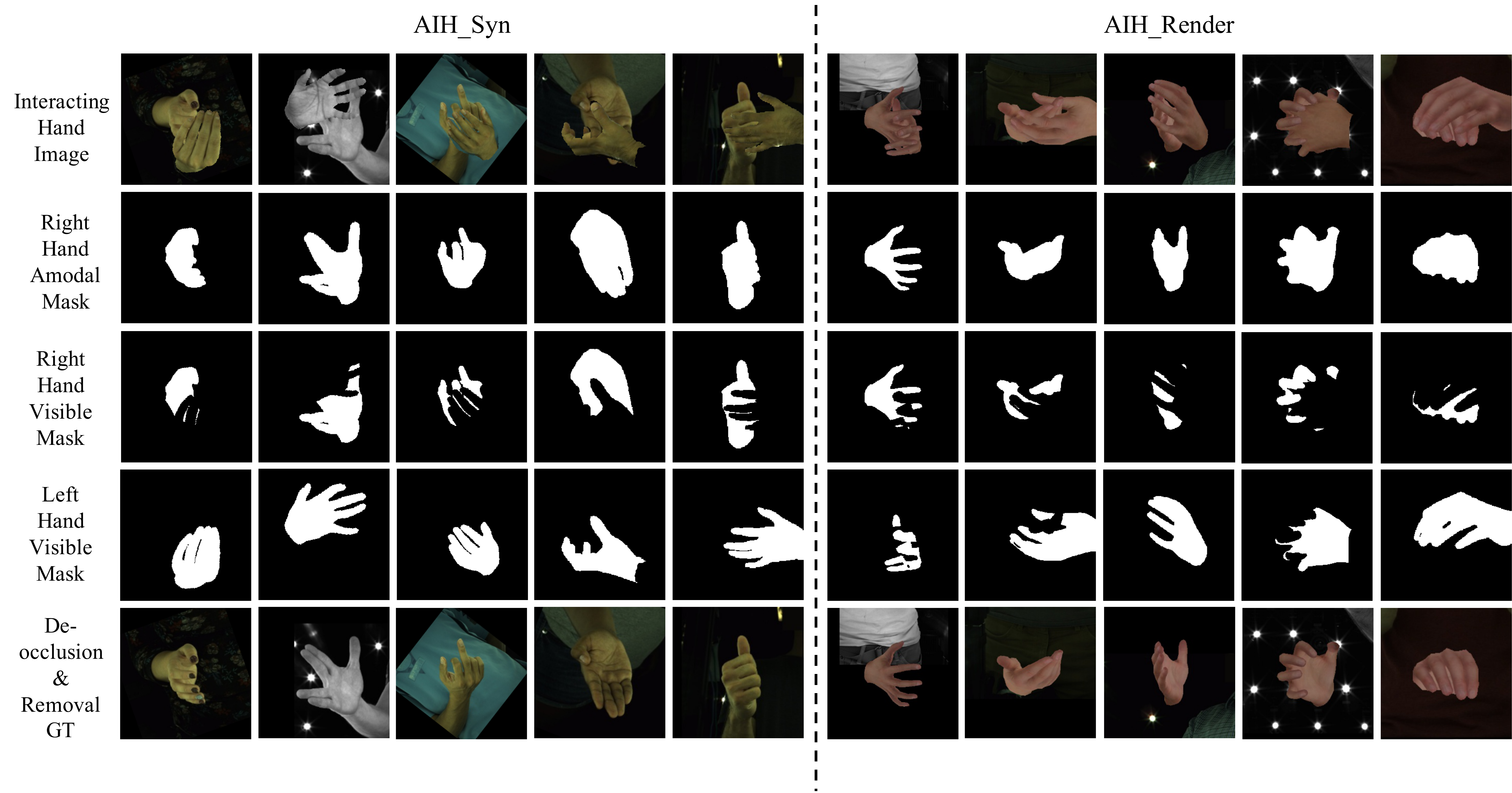}
\end{center}
   \caption{Visualization of our proposed Amodal InterHand (AIH) dataset. AIH\_Syn is obtained by simple 2D copy-and-paste, while AIH\_Render is generated by rendering the textured 3D interacting hand mesh to the image plane.
   }
\label{fig:AIH}
\end{figure*}

Our AIH dataset is constructed based on the well-known InterHand2.6M V1.0 dataset~\cite{moon2020interhand2}. As shown in Fig.~\ref{fig:AIH}, our proposed Amodal InterHand (AIH) dataset consists of two parts: AIH\_Syn and AIH\_Render. 
In total, AIH dataset consists of about 3 million images, where AIH\_Syn contains 2.2M samples and AIH\_Render contains over 0.7M samples. 
AIH\_Syn is generated by simple 2D image-level copy and paste, \ie copy the left single-hand image and paste it on the right single-hand image; AIH\_Render is obtained by rendering the textured interacting hand mesh to the image plane. Both AIH\_Syn and AIH\_Render contain the amodal and modal segmentation masks as well as the appearance content ground-truths.

\textbf{AIH\_Syn} We first get the hand mesh with the ground-truth MANO parameters of the single-hand samples from the InterHand2.6M V1.0 dataset, and then project it into the 2D image plane to get the amodal segmentation mask. Then, we filter out some bad samples in which MANO parameters or the corresponding image are not valid. As a result, we get over 250K cropped single-hand images with masks for each side hand. To generate the interacting two-hand samples, we randomly pick two hands with similar texture from both sides. Then we crop the left hand region given its amodal mask and paste it on the right hand image. Random scaling, rotation, and color jittering are applied to increase diversity.

\textbf{AIH\_Render} Although AIH\_Syn provides plenty of amodal data, such 2D level copy-and-paste can not generate mutual occlusion cases which are very common for interacting hands. Therefore, based on MANO parameters of InterHand2.6M V1.0 dataset, we decorate the corresponding hand mesh with random hand texture~\cite{HTML_eccv2020}, augment it with random translation and rotation, and finally render it to a random background image from the dataset. 

\section{Experiments}

\subsection{Implementation Details}
All experiments are conducted on 8 NVIDIA Tesla V100 GPUs. Training mini-batch size is set as 48 and Adam~\cite{kingma2014adam} is adopted for model parameter tuning.
\textbf{Details of HASM.} Our HASM is trained for 200k iterations with a learning rate of $2.5 \times 10^{-3}$ following the training setting of SegFormer~\cite{xie2021segformer}.
\textbf{Details of HDRM.} Our HDRM has an input resolution of $256 \times 256$. And the loss weights are set as $\lambda_1=0.1$, $\lambda_2=3.0$, $\lambda_3=0.1$, $\lambda_4=250.0$. We first train HDRM with ground-truth masks for 100k iteration, and then fine-tune it with segmentation masks for another 100k iteration. The learning rate of these two stages are $1.5 \times 10^{-3}$ and $1 \times 10^{-3}$ respectively.
\textbf{Details of SHPE.} Our SHPE has an input resolution of $256 \times 256$. We train the network for 300k iterations with an initial learning rate of $1 \times 10^{-3}$. The learning rate is decayed to $1 \times 10^{-4}$ and $1 \times 10^{-5}$ at the 100k and 200k iterations respectively. Other training settings are kept the same as those of MinimalHand~\cite{zhou2020monocular}.

\subsection{Datasets and Evaluation Metrics}

\textbf{Datasets.} The experiments are conducted on InterHand2.6M V1.0~\cite{moon2020interhand2} dataset and Tzionas dataset~\cite{tzionas2016capturing}.
\emph{InterHand2.6M V1.0 dataset}~\cite{moon2020interhand2} is a publicly available large-scale realistic pose estimation dataset for two-hand interactions. The dataset provides RGB images with semi-automatically annotated 3D poses, and MANO~\cite{romero2022embodied} parameters obtained from NeuralAnnot~\cite{moon2020neuralannot}. In the experiments, we follow the common practice~\cite{moon2020interhand2} to use the downsized $512\times334$ image resolution at 5 frames-per-second (FPS) version of the released dataset. Following the official configurations, the dataset is split into three branches, namely `H' for the human annotation branch, `M' for the machine annotation branch, and `ALL' for all data. The `M' branch data contains many unseen poses and more diverse sequences, which makes it more similar to real-world scenarios. Moreover, we notice that the `H' branch data contains missing or incomplete mesh annotations. In the experiments, we majorly conduct experiments on the `M' branch, but also report the results on `ALL' branch for comparisons. To focus on the interacting hands, the original dataset~\cite{moon2020interhand2} divides the whole dataset (IH26M-ALL) into single-hands subset (IH26M-SH) and interacting-hands subset (IH26M-IH). Following~\cite{rong2021ihmr}, we further select samples from the original ``IH26M-IH'' test set, and generate a more challenging subset called ``IH26M-Inter''. ``IH26M-Inter'' contains samples with more than 30 valid `ground-truth' 3D hand keypoints. Since InterHand2.6M~\cite{moon2020interhand2} is captured in a lab environment, its background diversity is relatively limited. To evaluate the generalization ability, we perform qualitative and quantitative experiments on the 
\emph{Tzionas dataset}~\cite{tzionas2016capturing}. Since Tzionas dataset does not provide a separate training set, we directly use it as the testing set to evaluate the model trained on the InterHand2.6M dataset~\cite{moon2020interhand2}. 

\textbf{Evaluation Metrics.} For InterHand2.6M~\cite{moon2020interhand2}, we report 3D Mean Per Joint Position Error (MPJPE). MPJPE is defined as the mean Euclidean distance between ground truth and predicted 3D joint locations, calculated after aligning the root joint for each left and right hand separately. The measurements are reported in millimeters (mm). For Tzionas dataset~\cite{tzionas2016capturing}, we follow the common practice~\cite{moon2020interhand2,boukhayma20193d,kim2021end} to use 2D end point error (EPE) for evaluation.

\subsection{Comparisons with state-of-the-art methods}

We compare with previous state-of-the-art pose estimation methods on the `ALL' branch and the `machine\_annot (M)' branch of InterHand2.6M V1.0 dataset~\cite{moon2020interhand2}. MPJPE (mm) is adopted to evaluate the 3D hand pose estimation accuracy. For fair comparisons, the AIH dataset is only used to train HASM and HDRM for amodal segmentation and de-occlusion. No pose annotations in AIH are used to train the pose estimator (SHPE). Table~\ref{tab:compare_with_sota} summarizes the experimental results. 

We first compare performances of three single-hand methods, \ie Boukhayma~\etal~\cite{boukhayma20193d}, Pose2Mesh~\cite{Choi_2020_ECCV_Pose2Mesh} and BiHand~\cite{yang2020bihand}. Our approach significantly outperforms all the state-of-the-art single-hand approaches. On the ``IH26M-ALL'' split, compared with BiHand~\cite{yang2020bihand}, our model reduces MPJPE from $25.10$mm to $10.97$mm, resulting in as much as $56\%$ error reduction. And in the more challenging ``IH26M-Inter'' split, our approach obtains about $47\%$ accuracy improvement. This shows existing single-hand pose estimators do not handle heavy hand-hand occlusions and are easily confused by the other distracting hand.

We also compare with recent two-hand pose estimation approaches, \ie InterNet~\cite{moon2020interhand2}, Rong~\etal~\cite{rong2021ihmr}, and DIGIT~\cite{fan2021digit}. We show superior performance over these 3D interacting hand pose estimation systems. For example, our approach significantly improves upon Moon~\etal~\cite{moon2020interhand2}’s state-of-the-art results from $14.21$mm to $10.97$mm (about 23\% error reduction) on the ``IH26M-ALL'' split. The clear performance gap validates the effectiveness of our framework. Overall, our approach consistently ranks the first across all evaluation protocols.

\begin{table}[tb]
\setlength\tabcolsep{3pt}
\caption{\textbf{Comparisons} with state-of-the-art methods on the `ALL' branch and the `machine\_annot (M)' branch of InterHand2.6M V1.0 Dataset. MPJPE (mm) is adopted to evaluate the 3D joint estimation accuracy.
The results marked with `*' are from~\cite{rong2021ihmr}.}
\centering
\scalebox{0.73}{
\begin{tabular}{c|cccc|ccc}
\hline
\multirow{2}{*}{Methods} & \multicolumn{4}{c|}{InterHand2.6M - ALL branch} & \multicolumn{3}{c}{InterHand2.6M - M branch} \\ \cline{2-8} 
 & \multicolumn{1}{c|}{IH26M-SH} & \multicolumn{1}{c|}{IH26M-IH} & \multicolumn{1}{c|}{IH26M-ALL} & IH26M-Inter & \multicolumn{1}{c|}{IH26M-SH} & \multicolumn{1}{c|}{IH26M-IH} & IH26M-ALL \\ \hline
*Boukhayma~\etal~\cite{boukhayma20193d} & \multicolumn{1}{c|}{\_} & \multicolumn{1}{c|}{\_} & \multicolumn{1}{c|}{27.14} & 31.46 & \multicolumn{1}{c|}{\_} & \multicolumn{1}{c|}{\_} & \_ \\  
*Pose2Mesh~\cite{Choi_2020_ECCV_Pose2Mesh} & \multicolumn{1}{c|}{\_} & \multicolumn{1}{c|}{\_} & \multicolumn{1}{c|}{27.10} & 32.11 & \multicolumn{1}{c|}{\_} & \multicolumn{1}{c|}{\_} & \_ \\  
*BiHand~\cite{yang2020bihand} & \multicolumn{1}{c|}{\_} & \multicolumn{1}{c|}{\_} & \multicolumn{1}{c|}{25.10} & 28.23 & \multicolumn{1}{c|}{\_} & \multicolumn{1}{c|}{\_} & \_ \\  
*Rong~\etal~\cite{rong2021ihmr} & \multicolumn{1}{c|}{\_} & \multicolumn{1}{c|}{\_} & \multicolumn{1}{c|}{17.12} & 20.66 & \multicolumn{1}{c|}{\_} & \multicolumn{1}{c|}{\_} & \_ \\  
DIGIT~\cite{fan2021digit} & \multicolumn{1}{c|}{\_} & \multicolumn{1}{c|}{14.27} & \multicolumn{1}{c|}{\_} & \_ & \multicolumn{1}{c|}{\_} & \multicolumn{1}{c|}{\_} & \_ \\  
InterNet~\cite{moon2020interhand2} & \multicolumn{1}{c|}{12.16} & \multicolumn{1}{c|}{16.02} & \multicolumn{1}{c|}{14.21} & 18.04 & \multicolumn{1}{c|}{12.52} & \multicolumn{1}{c|}{18.04} & 15.28 \\ \hline
\textbf{HDR (Ours)} & \multicolumn{1}{c|}{\textbf{8.51}} & \multicolumn{1}{c|}{\textbf{13.12}} & \multicolumn{1}{c|}{\textbf{10.97}} & \textbf{14.74} & \multicolumn{1}{c|}{\textbf{8.52}} & \multicolumn{1}{c|}{\textbf{14.98}} & \textbf{11.74} \\ \hline
\end{tabular}
}
\label{tab:compare_with_sota}
\end{table}

\begin{table}[h]
\caption{\textbf{Comparisons} with state-of-the-art methods on Tzionas dataset~\cite{tzionas2016capturing}. The results of other algorithms are from~\cite{kim2021end}. 2D EPE is adopted to evaluate pose results.}
    \resizebox{0.98\textwidth}{!}{
        \begin{tabular}{c|c|c|c|c||c|c}
            \hline 
            Model & Boukhayma~\etal~\cite{boukhayma20193d} & Wang~\etal~\cite{wang2020rgb2hands} & InterNet~\cite{moon2020interhand2} & Kim~\etal~\cite{kim2021end} & SHPE & SHPE+HDR \\
            \hline 
            EPE$\downarrow$ & 12.91 & 13.31 & 17.61 & 12.42 & 14.88 & \textbf{8.70} \\
            \hline
        \end{tabular}
	}
	\label{tab:tzionas}
\end{table}

We also follow~\cite{kim2021end} to report hand pose estimation results (EPE) on Tzionas dataset~\cite{tzionas2016capturing} in Table~\ref{tab:tzionas}. Our method (SHPE+HDR in the table) significantly improves upon the baseline SHPE, and outperforms the prior arts.

\subsection{Effect of Hand De-occlusion and Removal (HDR) Framework}

 As shown in Table~\ref{tab:TDRNet_M}, and Table~\ref{tab:TDRNet_ALL}, we conduct experiments on the `machine\_annot (M)' branch and the `ALL' branch of InterHand2.6M V1.0 dataset respectively. We compare the results with or without using our HDR framework. We notice that the recent state-of-the-art single-hand pose estimation (SHPE) method (MinimalHand~\cite{zhou2020monocular}) 
 struggles with occlusions and appearance ambiguity in interacting hand scenarios (IH26M-IH). To tackle these challenges, we propose a novel Hand De-occlusion and Removal (HDR) framework to perform hand de-occlusion and distractor removal. In Table~\ref{tab:TDRNet_M}, we show that our approach significantly improves upon the SHPE baseline in interacting hand scenarios, \eg from 40.98mm to 25.45mm (M, IH26M-IH). We find that adding AIH dataset for training will further improve the performance of SHPE, which validates the effect of AIH dataset. Experiments on the `ALL' branch have a similar phenomenon.

\begin{table}[tb]
\setlength\tabcolsep{5pt}
\caption{\textbf{Effect of HDR Framework.} Experiments are conducted on the `machine\_annot (M)' branch of InterHand2.6M V1.0 dataset. MPJPE (mm) is adopted to evaluate the 3D joint estimation accuracy.}
\centering
\scalebox{0.9}{
\begin{tabular}{l|c|c|c|c}
\hline
  \multirow{2}{*}{Methods} &  \multicolumn{2}{c|} {Train (M, IH26M-SH)} &
  \multicolumn{2}{c} {Train (M, IH26M-SH +AIH)} \\ \cline{2-5}
  & IH26M-IH & IH26M-ALL  & IH26M-IH & IH26M-ALL \\ \hline
  SHPE~\cite{zhou2020monocular}  & 40.98 & 25.78 & 32.27 & 21.66 \\ 
  +HDR (Ours)  & 25.45 & 17.98  & 24.59 & 17.80 \\ \hline
\end{tabular}
}
\label{tab:TDRNet_M}
\end{table}

\begin{table}[tb]
\setlength\tabcolsep{5pt}
\caption{\textbf{Effect of HDR Framework.} Experiments are conducted on the `ALL' branch of InterHand2.6M V1.0 dataset. MPJPE (mm) is adopted to evaluate the 3D joint estimation accuracy.}
\centering
\scalebox{0.9}{
\begin{tabular}{l|c|c|c|c}
\hline
  \multirow{2}{*}{Methods} &  \multicolumn{2}{c|} {Train (ALL, IH26M-SH)} &
  \multicolumn{2}{c} {Train (ALL, IH26M-SH +AIH)} \\ \cline{2-5}
  & IH26M-IH & IH26M-ALL & IH26M-IH & IH26M-ALL \\ \hline
  SHPE~\cite{zhou2020monocular} & 39.96 & 25.90  & 30.23 & 20.93 \\ 
  +HDR (Ours)  & 25.93 & 18.39  & 23.99 & 17.58 \\ \hline
\end{tabular}
}
\label{tab:TDRNet_ALL}
\end{table}

\subsection{Ablation Study}

\begin{table}[t]
\setlength\tabcolsep{4pt}
\caption{\textbf{Ablation Studies.} Experiments are conducted on the `machine\_annot (M)' branch of InterHand2.6M V1.0 dataset. We use MPJPE (mm) to evaluate the 3D joint estimation accuracy. $\Delta$ means the absolute (and relative) difference compared with our final model \#8. `w/o' is short for `without'.}
\centering
\scalebox{0.9}{
\begin{tabular}{l|c|c|c}
\hline
 & Methods &  MPJPE (mm) & $\Delta$ \\ \hline
 \#1 & SHPE~\cite{zhou2020monocular} only & 25.78 & +7.80 (43.4\%)  \\ 
 \#2 & w/o Removal &  24.16 & +6.18 (34.4\%) \\ 
 \#3 & w/o De-occlusion &  19.69 & +1.71 (9.5\%) \\ \hline
 \#4 & w/o Discriminator &  18.11 & +0.13 (0.7\%) \\
 \#5 & w/o Transformer Block &  18.85 & +0.87 (4.8\%) \\ \hline
 \#6 & AIH\_Render only &  18.10 & +0.12 (0.7\%) \\ 
 \#7 & AIH\_Syn only  &  18.35 & +0.37 (2.1\%) \\ \hline
 \#8 & Ours &  \textbf{17.98} & - \\ \hline
\end{tabular}
}
\label{tab:ablation}
\end{table}

In this section, we conduct ablation studies to evaluate the effectiveness of the key components of our approach on the ‘machine annot (M)’ branch of InterHand2.6M V1.0 dataset~\cite{moon2020interhand2}. For fair comparisons, in all the ablation experiments, we use the same SHPE~\cite{zhou2020monocular} trained on the IH26M-SH set. 

\textbf{Analysis of Hand De-occlusion and Removal (HDR) Module.} 
There are two major challenges of hand pose estimation in interacting scenarios, \ie severe self-occlusion, and ambiguity caused by the homogeneous appearance of hands. As shown in Table~\ref{tab:ablation}, \#1, \#2, \#3, and \#8, we conduct ablative experiments to quantitatively evaluate the effect of De-occlusion and Removal. Comparing \#2 and \#8, we observe that disabling `Removal' will dramatically increase the MPJPE by 34.4\%. Comparing \#3 and \#8, we see that disabling `De-occlusion' increases the MPJPE by 9.5\%. If we only apply SHPE~\cite{zhou2020monocular} without HDR, the errors are further increased. These results clearly show that (1) state-of-the-art SHPE~\cite{zhou2020monocular} is sensitive to self-occlusions and inter-hand ambiguities (2) HDRM is effective in handling the aforementioned two major challenges. 

\textbf{Analysis of Model Design Choices.}
We empirically validate the model design choice of HDRNet, especially \emph{Discriminator} and the \emph{Transformer} block. Discriminator is applied to enhance the image recovery quality by adversarial training. Comparing \#4 and \#8 in Table~\ref{tab:ablation}, we observe that although Discriminator helps in improving the quality of the recovered image, its influence on the final results is only marginal (0.7\%). The Transformer block enhances image feature interactions, enlarges the receptive fields, and focuses more on important image regions. Comparing \#5 and \#8 in Table~\ref{tab:ablation}, we see that without using the Transformer block impacts the final results by a clear margin (4.8\%).

\textbf{Analysis of AIH\_Syn and AIH\_Render.} Our proposed AIH dataset is composed of two subsets, namely AIH\_Syn and AIH\_Render. Both have their own advantages and disadvantages. AIH\_Syn retains more detailed and realistic appearance features, while AIH\_Render considers the inter-dependencies between two hands to avoid physically implausible
configurations. Using a combination of these two sets to train the HDRNet will achieve the best performance. In Table~\ref{tab:ablation}, comparing \#6, \#7, and \#8, we compare different training settings for HDRNet. We notice that it already achieves reasonably good results even if we only use one of the two sets. For example, using ``AIH\_Render only'' to train HDRNet, we can achieve $18.10$ MPJPE (mm), which is only marginally worse than the final model \#8. We also empirically find that ``AIH\_Render'' seems to have a larger impact on the final results than ``AIH\_Syn'' does. 

\subsection{Time Complexity Analysis}
We analyze the time cost on one Tesla P40 GPU in a single thread. On average, HASM, HDRM, and SHPE take 12.6 ms, 0.6 ms, and 34.0 ms per frame (including two hands) respectively. The time cost of HDRM (our major contribution) is only a small proportion of the total time cost (0.6 vs 47.2 ms). 

\subsection{Qualitative Results}

\begin{figure*}[t]
\begin{center}
   \includegraphics[width=0.99\linewidth]{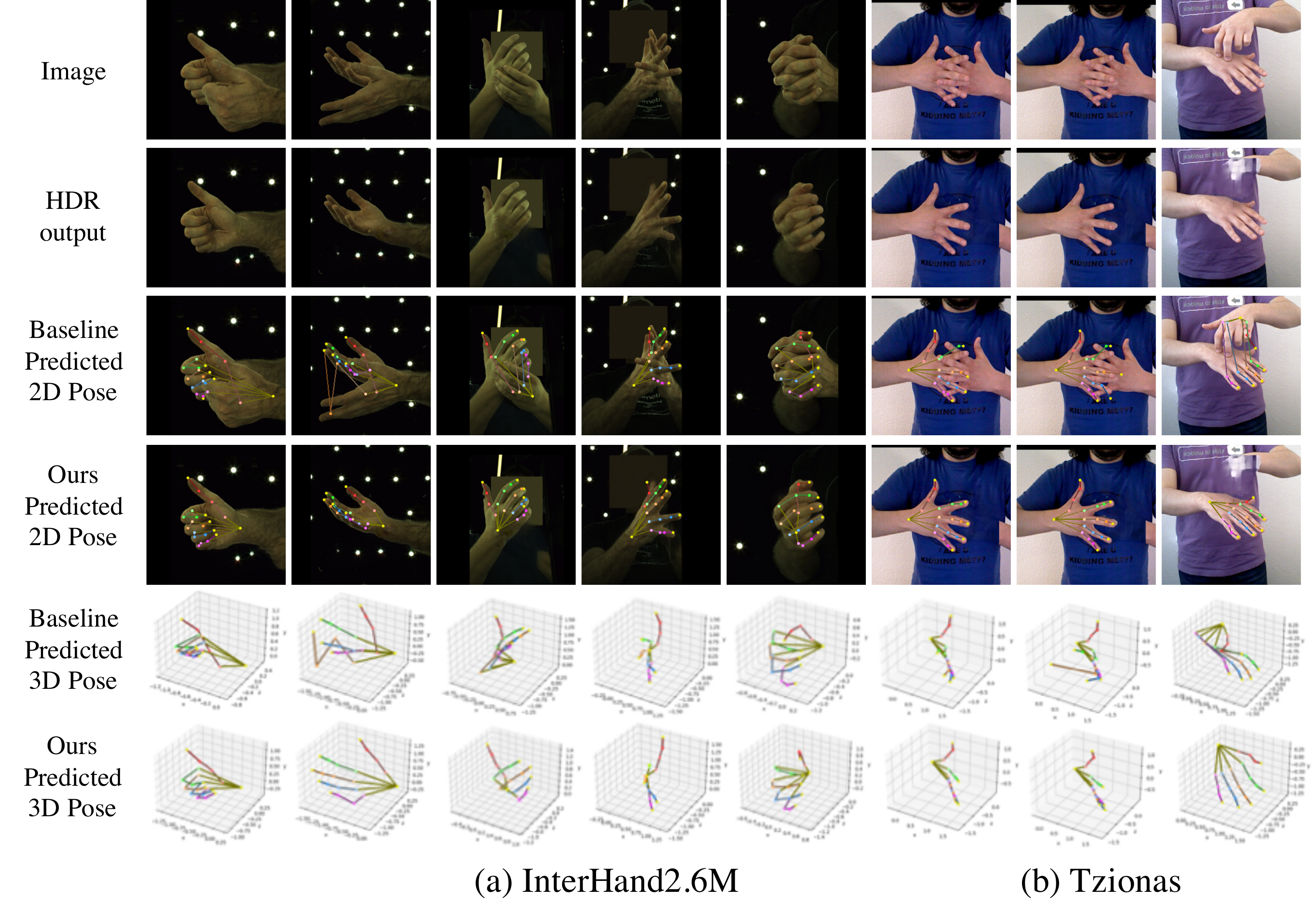}
\end{center}
   \caption{Qualitive results of how HDR helps in handling severe hand-hand occlusion and appearance ambiguities of two hands. Best viewed in color. 
   }
\label{fig:qualitative}
\end{figure*}

In Fig.~\ref{fig:qualitative}, we provide qualitative analysis on InterHand2.6M~\cite{moon2020interhand2} and Tzionas dataset~\cite{tzionas2016capturing} to illustrate how HDR helps in handling severe hand-hand occlusion and the homogeneous appearance of hands. We see that HDR recovers the appearance in the occluded region and removes the distractor in challenging hand-hand occlusion cases.
The results on Tzionas dataset~\cite{tzionas2016capturing} further validates the generalization ability of our proposed framework.

\section{Conclusions and Limitations}

Interacting hand pose estimation is important but challenging due to severe hand-to-hand occlusion and ambiguity caused by the other distracting hand. In this paper, we propose to decompose the task into two relatively simple sub-tasks, \ie (1) Hand De-occlusion and Removal (HDR), (2) Single Hand Pose Estimation (SHPE). Through HDR, we can simplify the case, which an off-the-shelf SHPE can handle. We empirically verified the effectiveness of our HDR framework on the InterHand2.6M and Tzionas dataset. 
Our limitations mainly lie in artifacts produced by HDRM. Improving the image recovery quality requires efforts in various research fields, which we will explore in the future. 

\textbf{Acknowledgement.} 
We would like to thank Wentao Jiang, Wang Zeng, Neng Qian, Yumeng Hu, Lixin Yang, Yu Rong, Qiang Zhou and Jiayi Wang for their helpful discussions and feedback.
Mengxiang Lin is supported by State Key Laboratory of Software Development Environment under Grant No SKLSDE 2022ZX-06. Ping Luo is supported by the General Research Fund of HK No.27208720, No.17212120, and No.17200622.
Wanli Ouyang is supported by the Australian Research Council Grant DP200103223, Australian Medical Research Future Fund MRFAI000085, CRC-P Smart Material Recovery Facility (SMRF) – Curby Soft Plastics, and CRC-P ARIA - Bionic Visual-Spatial Prosthesis for the Blind.

\clearpage
% ---- Bibliography ----
%
% BibTeX users should specify bibliography style 'splncs04'.
% References will then be sorted and formatted in the correct style.
%
\bibliographystyle{splncs04}
\bibliography{egbib}

\clearpage

\appendix
\section*{\Large Appendix}
\setcounter{table}{0}
\renewcommand{\thetable}{A\arabic{table}}
\setcounter{figure}{0}
\renewcommand{\thefigure}{A\arabic{figure}}

\section{Video Demo}

To justify the generalization ability and the potential of our proposed method in real-world applications, we run our approach on several video clips from the Tzionas dataset~\cite{tzionas2016capturing}. Note that our models are only trained on InterHand2.6M V1.0 dataset~\cite{moon2020interhand2} and our proposed AIH dataset. Tzionas dataset~\cite{tzionas2016capturing} is \emph{unseen} during training. 

In the video demo\footnote{Our video demo can be downloaded from \url{https://connecthkuhk-my.sharepoint.com/:v:/g/personal/js20_connect_hku_hk/EW_S3kZu97xPlMk_HQLAJVMBtizU48sGh4jXwvUuyugFRw?e=u2GB6I}.}, we compare our approach with `Baseline' which is a Single-Hand Pose Estimator (SHPE). For fair comparisons, both `Ours' and `Baseline' employ the same SHPE~\cite{zhou2020monocular} trained on the ‘ALL’ branch of InterHand2.6M V1.0 dataset~\cite{moon2020interhand2}. Note that this model is the same as the one used in Sec. 5.6 (Fig. 5) of the main paper.
The pose results are directly obtained from the output of the SHPE model without temporal smoothing.

We first visualize the predicted amodal/visible mask of both hands. Given the segmentation mask, we obtain the corresponding single-hand box (`red' for the right hand, and `green' for the left hand). We also demonstrate the results of Hand De-occlusion and Removal Module (HDRM). In order to tackle the severe hand-hand occlusion problem, HDRM applies the hand de-occlusion technique to recover the appearance (texture) in the occluded region. In the meanwhile, it also removes (inpaints) the distracting hand to handle the ambiguity caused by the homogeneous appearance of hands. Finally, our approach obtains better 3d hand pose estimation results. 

Although the quality of the image recovery is satisfactory in most cases, there are still some problems in some difficult situations. For example, the boundary of the hand segmentation can be over-smoothed, leading to undesirable artifacts around the hand. This problem can be mitigated by applying an advanced amodal/visible mask segmentation model, which we will explore in the future.

\section{More Examples of AIH Dataset}

\begin{figure}[t]
\begin{center}
%\fbox{\rule{0pt}{2in} \rule{0.9\linewidth}{0pt}}
   \includegraphics[width=0.99\linewidth]{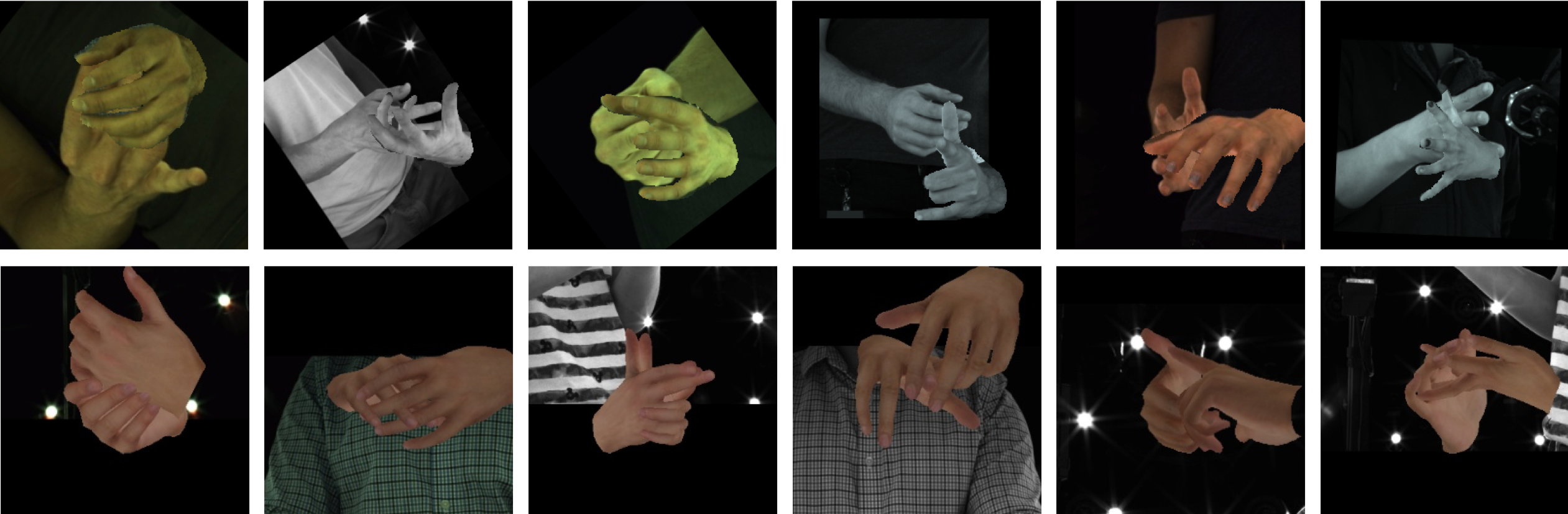}
\end{center}
   \caption{\textbf{Top:} More examples of our proposed AIH\_Syn dataset. \textbf{Bottom:} More examples of our proposed AIH\_Render dataset.}
\label{fig:AIH_supp}
\end{figure}

In this section, we present more examples of our proposed Amodal InterHand (AIH) dataset. Our AIH dataset consists of two parts: AIH\_Syn and AIH\_Render. AIH\_Syn is constructed by copy-and-paste while AIH\_Render is constructed by rendering the textured interacting hand mesh to the image plane. As shown in Fig~\ref{fig:AIH_supp}, both AIH\_Syn and AIH\_Render have great diversity in hand poses, textures, occlusion and interaction types.

\end{document}